\documentclass{article}

\usepackage{arxiv}

\usepackage[utf8]{inputenc} 
\usepackage[T1]{fontenc}    
\usepackage{hyperref}       
\usepackage{url}            
\usepackage{booktabs}       
\usepackage{amsfonts}       
\usepackage{nicefrac}       
\usepackage{microtype}      
\usepackage{lipsum}
\usepackage{graphicx}

\usepackage{times}
\usepackage{epsfig}
\usepackage{graphicx}
\usepackage{amsmath}
\usepackage{amssymb}
\graphicspath{ {./images/} }

\usepackage[linesnumbered,ruled,lined, noend]{algorithm2e}
\SetKwRepeat{Do}{do}{while}
\usepackage[flushleft]{threeparttable}
\usepackage{floatrow}
\usepackage{enumitem}
\newfloatcommand{capbtabbox}{table}[][\FBwidth]
\usepackage{soul}

\usepackage{mathtools}

\usepackage{multirow}
\usepackage{multicol}

\usepackage{amsmath}

\usepackage{enumitem}

\newcommand{\comment}[1]{}

\title{Learnable Discrete Wavelet Pooling (LDW-Pooling) for Convolutional Networks}

\author{
Bor-Shiun Wang \\
    College of Artificial Intelligence and Green Energy\\
    National Yang Ming Chiao Tung University\\
    \texttt{eddiewang.ai08@nycu.edu.tw} \\
    \And
Jun-Wei Hsieh \\
  College of Artificial Intelligence and Green Energy\\
  National Yang Ming Chiao Tung University\\
  \texttt{jwhsieh@nctu.edu.tw} \\
   \And
Ming-Ching Chang \\
  University at Albany - SUNY\\
  \texttt{mchang2@albany.edu} \\
  \And
Ping-Yang Chen \\
  Department of Computer Science\\
  National Yang Ming Chiao Tung University\\
  \texttt{pingyang.cs08g@nctu.edu.tw} \\
  \And
Lipeng Ke \\
    University at Buffalo \\
    \texttt{lipengke@buffalo.edu} \\
    \And
Siwei Lyu \\
    University at Buffalo \\
    \texttt{siweilyu@buffalo.edu} \\
}

\begin{document}
\maketitle
\begin{abstract}
Pooling is a simple but widely used layer in modern deep CNN architectures for feature aggregation and extraction. Typical CNN design focuses on the conv layers and activation functions, while leaving the pooling layers with fewer options. We introduce the Learning Discrete Wavelet Pooling (LDW-Pooling) that can be applied universally to replace standard pooling operations to better extract features with improved accuracy and efficiency. Motivated from the wavelet theory, we adopt the low-pass (L) and high-pass (H) filters horizontally and vertically for pooling on a 2D feature map. Feature signals are decomposed into four (LL, LH, HL, HH) subbands to retain features better and avoid information dropping. The wavelet transform ensures features after pooling can be fully preserved and recovered. We next adopt an energy-based attention learning to fine-select crucial and representative features. LDW-Pooling is effective and efficient when compared with other state-of-the-art pooling techniques such as WaveletPooling and LiftPooling. Extensive experimental validation shows that LDW-Pooling can be applied to a wide range of standard CNN architectures and consistently outperform standard (max, mean, mixed, and stochastic) pooling operations.
\end{abstract}


\section{Introduction}
Convolutional Neural Networks (CNNs) have been widely used in many domains, including object detection, segmentation, and recognition.  In the design of CNN architecture, downsampling layers with pooling and stride-convolutions are common operations that can effectively aggregate features. Standard pooling methods such as simple average pooling, max/min pooling are widely used for multiple purposes in CNN, to (1) fine-select important features, (2) aggregate spatially local features, and (3) retain important features for efficient processing in subsequent layers. Although being widely used, there are several drawbacks of the simple pooling methods. First, over-aggressive filtering inevitably results in information loss and degraded performance. 
Second, the simple reduction of spatial resolution breaks {\em shift-invariance}~\cite{Making:ICML2019}. 
Third, pooling operations are generally not {\em invertible} --- upsampling the down-sampled feature maps cannot recover the lost information. Finally, feature reduction can be pushed to the extreme, that the feature vector of a small object can reduce into a single-pixel or disappear in the deeper layers, thus hinders accurate discrimination. To this end, many recent designs of more sophisticated but effective pooling methods are proposed~\cite{LiftPooling:ICLR2021,WaveletPooling:ICLR2018}. However, due to the broad applicability, the potential of a better pooling mechanism is yet to be thoroughly explored. This paper investigates a bidirectional pooling mechanism that can preserve details when down-sampling the feature maps and retain finer spatial information to recover detailed up-sampled feature maps. 

Max pooling is the most widely used downsampling operation in popular CNN architectures, where the strongest activation within each neighborhood is kept in the filtering scope. However, such strongest activations may not always be the best choice for pooling. In \cite{Stochastic:ICLR2013}, a stochastic pooling technique is proposed to ensure that non-maximal activations can also be utilized. To address the shift-invariance problem, an anti-aliasing filter with a smoothing kernel is adopted in \cite{Making:ICML2019} to improve ImageNet classification. In general, excessive pooling can degrade object detection or recognition accuracy due to the inability to preserve discriminative features or details.  In \cite{Lip:ICCV2019}, the Local Importance-based Pooling (LIP) outperformed hand-crafted pooling layers by a large margin, with a penalty of lower efficiency. In \cite{Adaptive:WACV2020}, an adaptive pooling technique can adjust the level of details to preserve. However, this adaptive technique needs a parallel implementation for efficiency consideration.  

\begin{figure}[t]
\centerline{
  {\footnotesize (a)} \includegraphics[width=0.45\linewidth]{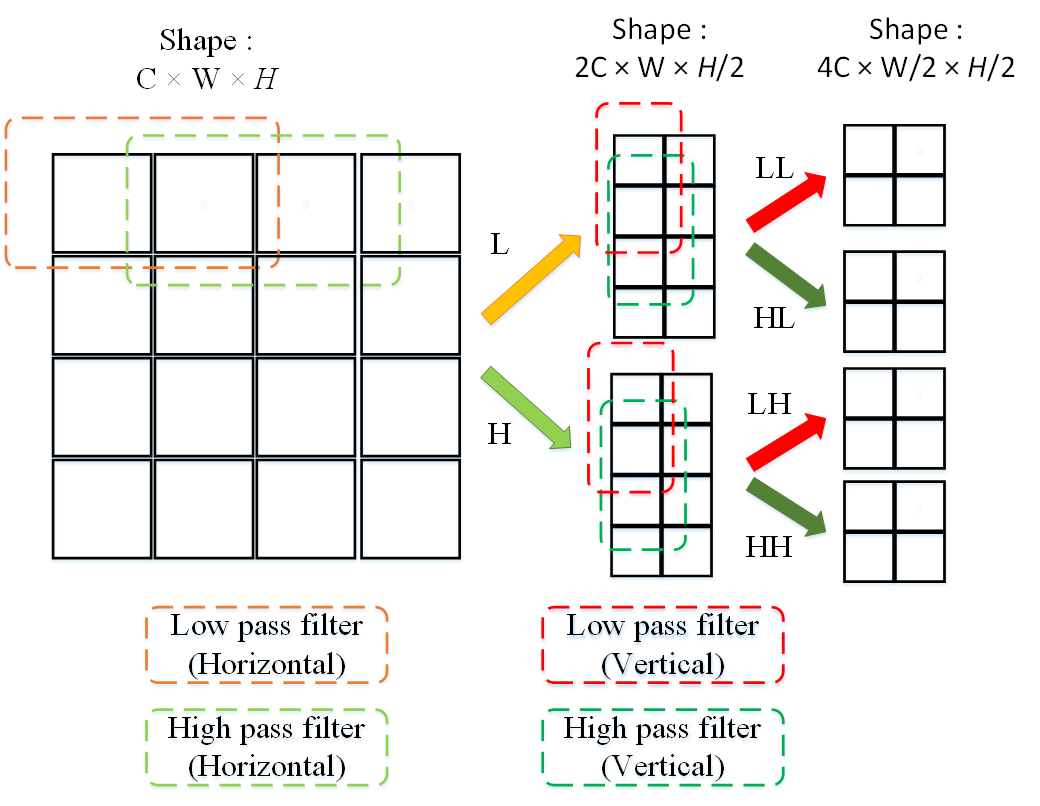}
  {\footnotesize (b)} 
  \includegraphics[width=0.45\linewidth]{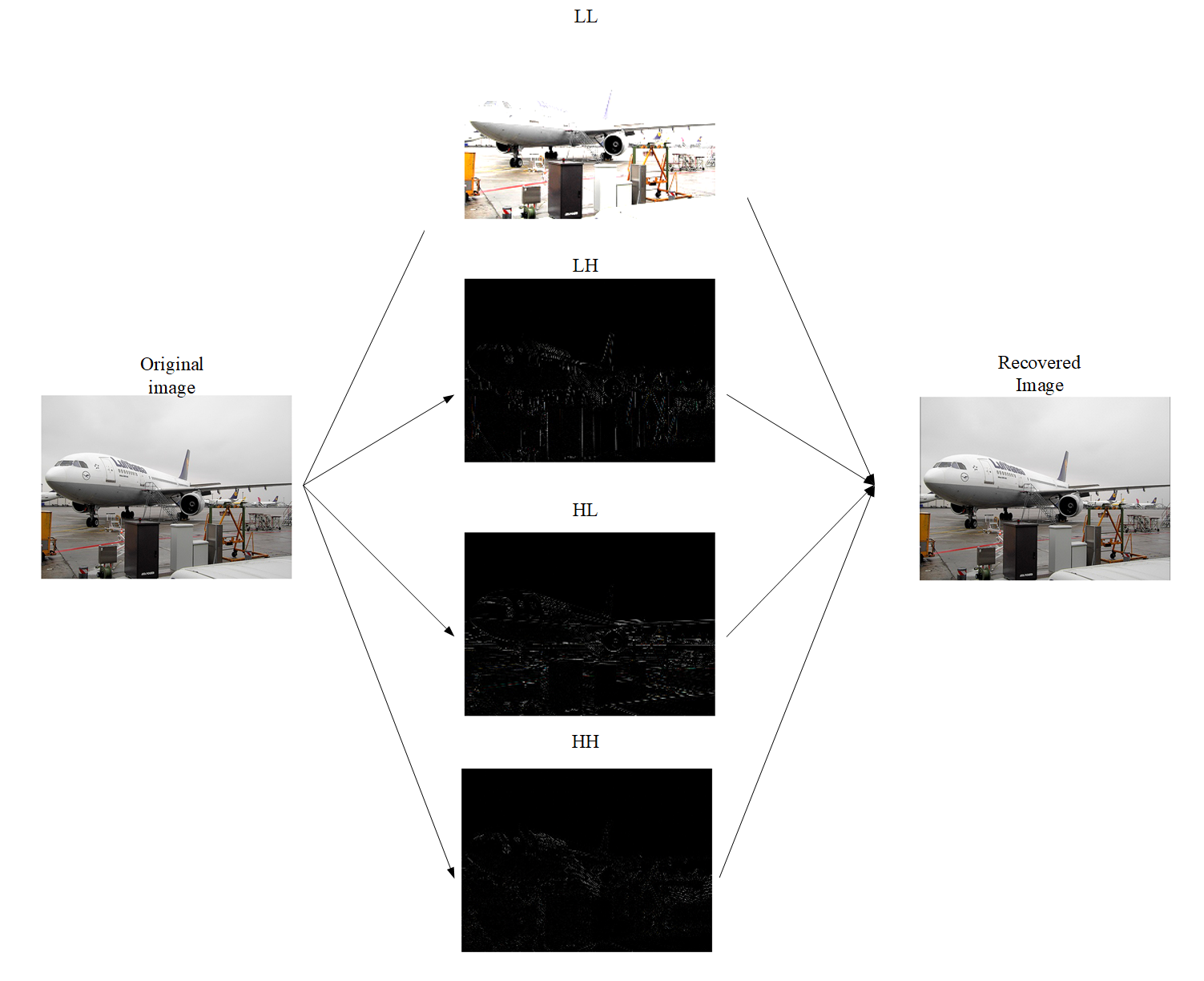}
}
\caption{
{\bf Overview of the proposed LDW-Pooling method.} 
(a) Given an input signal of $C$ channels of width $W$ $\times$ height $H$, we use high-pass (H) and low-pass (L) filters in the size of $1\times K$ (horizontally) and $K\times1$ (vertically) to decompose each 2D feature map into four subbands, LL, HL, LH, HH. 
(b) Reconstructed result after LDW-Pooling decomposition and recovering by UNet with PSNR=33.1}.
\label{fig:LDWpool}
\end{figure}

In this work, we propose a {\em learnable} discrete wavelet (LDW) pooling technique to better aggregate local features, such that lost information can be better recovered. 
Moreover, in LDW-Pooling, we decompose 2-D convolutions into the $1\times K$ and $K\times 1$ kernels to efficiently extract horizontal and vertical features with a learnable attention mechanism. Figure~\ref{fig:LDWpool} shows the overview of the proposed LDW-Pooling method.

The main contributions of this paper are summarized in the following:  
\begin{itemize}
\item We propose a new LDW-Pooling scheme by decomposing the input signal into four discrete wavelet subands (LL, LH, HL, HH) without information loss. With a new learnable energy attention design, LDW-Pooling enables accurate and fast pooling that can be combined and used in mainstream CNN applications.

\item The {\em reversibility} of LDW-Pooling is well-suited for fine-grained object classification for objects in both small and large sizes with higher accuracy.
 
\item The LDW-Pooling performs 2-D convolution by decomposing it into two 2 runs of 1-D convolutions. It is thus more efficient than traditional pooling methods. It can be easily integrated with different backbone thanks to this efficient and reversible design.
        
\item LDW-Pooling outperform the state-of-the-art classification method on CIFAR-10, CIFAR-100, and ImageNet datasets in terms of accuracy and efficiency. Results also show great generalization ability on various backbones.
    
\end{itemize}


\section{Background}
\label{sec:related}

{\bf Pooling in CNN.}
In CNN architectures, pooling is a widely used technique to improve the feature extraction efficiency and robustness to input variations. Two most popular pooling methods are average pooling and max pooling. The former is equivalent to blurred-downsampling and the latter is generally more effective and popularly used in many CNN backbones. Although being simple and effective, these standard pooling methods lose details in downsampling~\cite{Stochastic:ICLR2013,Mixedpooling:Springer2014}.  To maintain structured details, the detail-preserving pooling (DPP) in \cite{detailpreserving:CVPR2018} can magnify spatial changes and thus preserves some structural details.

{\bf Learning based pooling.}
To make the pooling learnable, \cite{Generalizing:AISTATS2016} learns adapted pooling on complex and variable patterns via a linear combination of max and average pooling. 
A stochastic pooling was proposed in \cite{Stochastic:ICLR2013} for solving the over-fitting problem in CNN architectures by using a multinomial distribution to aggregate activations within a region.  Similarly, in \cite{s3pool:CVPR2017}, S3 pooling was proposed to embed randomness into a multinomial probabilistic model, to select local activations with proper weights. 
A Gaussian model is modified in \cite{Gaussian:nips2019a} for probabilistic pooling with trainable parameters, where local activations can be aggregated based on their local statistics. 
Toward trainable pooling operation, a global feature guided flexible pooling in \cite{Gobal:ICCV2019} selects the pooling function based on the maximum entropy principle.  
 
{\bf Wavelet related pooling methods.} It is known that standard (max and average) poolings can drop details in downsampling, especially for important details that are with less intensity than the insignificant ones~\cite{Mixedpooling:Springer2014}. Several pooling approaches tried to preserve details by incorporate wavelet representations into CNNs. A wavelet pooling algorithm in \cite{WaveletPooling:ICLR2018} decomposes input signals to different subbands using wavelet transform, such that the feature contents can more accurately represented with fewer artifacts. The first layer of ResNet is modified in \cite{Scaling:ICCV2017} with a wavelet scattering network to maintain performances using a smaller number of pooling parameters. In \cite{WaveletConv:CoRR2018}, Haar wavelet CNNs are combined with a multi-resolution analysis for texture classification and image annotation.  

{\bf Feature-preserving and invertible pooling.}
The Lifting Scheme within the CNN networks in \cite{Adaptive:WACV2020} improves texture and object classification with less parameters. Their framework focuses on building an interpretable network by integrating multiresolution analysis, rather than pooling.  
The BlurPool with a smoothing kernel in \cite{Making:ICML2019} can reduce aliasing and improve ImageNet classification accuracy, however the pooling still results in detail lost. 
In the above wavelet-based pooling techniques or multi-resolution analysis, Haar wavelets are used for band decomposition. Thus in these methods, the wavelet filter are handcrafted, with fixed values, and thus not adaptive to handle data-dependent tasks. 
These pooling methods not invertible (that is, the lost information in the downsampling process cannot be recovered). Inspired by the Lifting Scheme~\cite{Liftingscheme:SIAM1998}, LiftPooling~\cite{LiftPooling:ICLR2021} allows the CNN architecture to be invertible and thus can achieve better robustness against input corruptions and perturbations. The LiftDownPool module was created to decompose the input into four sub-bands with the Haar wavelet. Next, another LiftUpPool module performs upsampling from these four sub-bands without losing information. The parameter space of LiftPooling is high, since all subbands must be used to achieve the best performance. Similar to other wavelet-based pooling methods, the Haar wavelet filters used for generating the four bands (LL, LH, HL, and HH) are handcrafted and not learnable from data.  

Compared to all surveyed methods, in this paper, we propose a learnable pooling scheme based on discrete wavelets that is {\em invertible, adaptive, and efficient}. The proposed LDW-Pooling learns the wavelet filter parameters from data for subband decomposition and recovery. Since the filters are learned from image contents, it is adaptive and potentially perform better for image classification and recognition. With a new energy attention learning, our method enables accurate and fast pooling that is suitable for CNN classification of objects of various sizes. It is computational efficient due to the decomposition of the 2-D convolution into two 1-D convolution operations.         

\section{Methods}
\label{sec:method}

\subsection{Discrete Wavelet Pooling}
\label{sec:dw:pool}

We employ the learnable discrete wavelet pooling to extract features and better retain them with less computation.
Let $X \in {R^{H \times W }}$ denote an input feature map.  The local pooling operation on $X$ can be defined as:
\begin{equation}
Y_{(x',y')} = \sum_{(\Delta x_i,\Delta y_i) \in \Omega} \; w_i \; X_{(x+ \Delta x_i,y+\Delta y_i)} \qquad s.t. \quad \sum_{i} \left(w_i\right)^2 =1, \; w_i \ge 0,
\label{eq:GeneralPooling}
\end{equation}
where $\Omega$ indicates a kernel region, and $w_i$ indicates the weighting parameter at each position. The division $(\frac{x}{x'}, \frac{y}{y'}$) stands for the stride factor, $i.e.$,  $x=2x'$ and $y=2y'$ for $2 \times 2$ stride.  For the case of average pooling, $Y$ represents the low-frequency part of $X$; and the high-frequency part of $X$ is discarded (in other words, the details of $X$ are missing after pooling).  
Max pooling keeps the maximal activation and thus performs generally better than average pooling on feature selection. However all non-maximum signals are discarded after max pooling. 
The widely used down-sampling scheme of {\em convolution with stride 2} in CNN also results in information loss. 
An intuitive motivation of our work is to develop a pooling operation that can extract and retain both low and high frequency parts of $X$. 
Inspired by the wavelet theory and the strength of invertability of wavelet transform, our wavelet based pooling is invertible in nature. With the new design of learning wavelet parameters from data, our LDW-Pooling can excel in feature-preserving and computational efficiency.  

Let $W^L$ and $W^H$ denote the low-pass and high-pass wavelet kernel weights, respectively. According to the property of wavelet transform,  $W^H$ and $W^L$ should satisfy the constraints:
\begin{equation}
\label{eq:FilterConstraints}
\sum_{i \in {\bf K}} \left(W^L(i)\right)^2 = 1 \quad \text{and} \quad \sum_{i\in {\bf K}} W^H(i) = 0,
\end{equation}
where ${\bf K}$ represents a 1-D kernel of size $K = |{\bf K}|$.  As shown in Figure~\ref{fig:LDWpool}, the two horizontal poolings are first performed using $W^L$ and $W^H$ respectively, and then two vertical pooling are performed subsequently. The horizontal pooling decomposes the incoming map along the $X$-direction to low-frequency signal $Y_{(x',y)}^L$ and high-frequency signal $Y_{(x',y)}^H$: 
\begin{multicols}{2}
\begin{equation}
\label{eq:LowPassPoolinginX}
    Y_{(x',y)}^L = \sum_{i \in {\bf K}} W^L(i) \; X_{(x+i,y)},
\end{equation}
\begin{equation}
\label{eq:HighPassPoolinginX}
    Y_{(x',y)}^H = \sum_{i \in {\bf K}} W^H(i) \; X_{(x+i,y)}.
\end{equation}
\end{multicols}

%
Assume there are $C$ channels of input feature maps. After the horizontal pooling, there are $2C$ channels of resulting feature maps with half sizes along the $X$ direction. 

Next, two vertical pooling operations continue on $Y_{(x',y)}^L$ and $Y_{(x',y)}^H$, respectively.  Regarding $Y_{(x',y)}^L$, we obtain two subbands:

\begin{multicols}{2}
\begin{equation}
\label{eq:LLPooling}
    Y_{(x',y')}^{LL} = \sum_{j \in {\bf K}} W^L(j) \; Y^L_{(x',y+j)},
\end{equation}
\begin{equation}
\label{eq:LHPooling}
    Y_{(x',y')}^{LH} = \sum_{j \in {\bf K}} W^H(j) \; Y^L_{(x',y+j)}.
\end{equation}
\end{multicols}
In addition, for $Y_{(x',y)}^H$, we obtain the other two subbands $Y_{(x',y')}^{HL}$ and  $Y_{(x',y')}^{HH}$ are obtained:

\begin{multicols}{2}
\begin{equation}
\label{eq:HLPooling}
    Y_{(x',y')}^{HL} = \sum_{j \in {\bf K}} W^L(j) \; Y^H_{(x',y+j)}.
\end{equation}
\begin{equation}
\label{eq:HHPooling}
    Y_{(x',y')}^{HH} = \sum_{j \in {\bf K}} W^H(j) \; Y^H_{(x',y+j)}.
\end{equation}
\end{multicols}

Note that only 1-D convolutions are used to obtain the pooling results $Y_{(x',y')}^{LL}$, $Y_{(x',y')}^{LH}$, $Y_{(x',y')}^{HL}$, and $Y_{(x',y')}^{HH}$ as well.  The time complexity for performing Eqs.\eqref{eq:LowPassPoolinginX}-\eqref{eq:HHPooling} is ${\cal O} (KWH)$.
In comparison, standard pooling techniques such max pooling or average pooling on conv stride 2 are all based on a 2-D kernel.
Their complexity is ${\cal O} (K^2WH)$, and the output signal is not invertible. 

For the best of our knowledge, all other existing wavelet-based pooling techniques~\cite{LiftPooling:ICLR2021,WaveletPooling:ICLR2018,Adaptive:WACV2020,Scaling:ICCV2017,WaveletConv:CoRR2018} perform signal decomposition based on a fixed Haar wavelets design: $W^L = \{ \frac{1}{2}, \frac{1}{2} \}$ and $W^H=\{\frac{1}{2}, \frac{-1}{2}\}$. Such handcrafted wavelet parameters are not learnable nor adaptive to application tasks.
Another drawback of these methods are that, after each pooling operation, the number of feature channels increases to become quadruple. Their parameter spaces are higher, since all subbands shall be used to achieve the desired CNN performance.  
In comparison to these method, our proposed LWD-Pooling adopts a learnable scheme to learn filter parameters from data for subband decomposition. Such learned 1-D wavelet parameters can better recover pooled signals with improved efficacy and efficiency, and the total parameters are greatly reduced.    


\subsection{Learning Discrete Wavelet Pooling Filter Parameters}
\label{sec:learn:param}

We next describe the design of the loss function for data-driven learning of discrete wavelet (DW) filter parameters. To prevent the filter weights from approaching mostly zero during the training process, we enforce the following constraints in the loss design.

For the low-pass DW filter, the square sum of filter weights should be equal to $1$ according to Eq.~\eqref{eq:FilterConstraints}.  Since the filters are to be learned from data, the values of the low-pass DW filter can be distinct to gain better performance. And we empirically confirmed such proposition. Let $W^L$ denotes the low-pass DW filter weights. According to this constraint from Eq.~\eqref{eq:FilterConstraints} during training, we introduce the following loss term for the low-pass DW filters:
\begin{equation}
\label{equ:con1}
L_{Low} = \left[\sum_i^K W^L(i) ^ 2 - 1\right] ^ 2 + \left[\sum_i^K \left( W^L(i)\right) - \sqrt2 \right] ^ 2.
\end{equation}
This constraint minimizes the squared differences between the sum of low-pass wavelet weights and $\sqrt{2}$.  For the high-pass DW filters, similarly, direct summation of high-pass filter weights in training may result in all zeros (as a trivial solution that is not desirable). Since the sum of high-pass filter weights must be zero, we introduce the following constraint $L_{High}$ in the loss term for the high-pass DW filters:
\begin{equation}
\label{equ:HighPassConstraint}
L_{High} = \left( \sum_i^K{W^H(i)} ^ 2 - 1 \right) ^ 2 
+\left( \sum_i^K{W^H(i)} \right) ^ 2,
\end{equation}
where $W^H$ denote the high-pass filter weight. The first term of Eq.~\eqref{equ:HighPassConstraint} enforce $W^H$ to become unit vectors.  

To retain the {\em reversibility} of DW pooling, the energy of $W^L$ and $W^H$ should be equal to 1.  Thus, we introduce the following reversibility loss term:
\begin{equation}
\label{equ:con3}
L_{Reverse} = \left[\left(\sum_i^K{W^L(i)} ^ 2
+
\sum_i^K{W^H(i)} ^ 2\right) - 2\right] ^ 2.
\end{equation}

Finally, to make the symmetrical filters, we introduce the following symmetrical loss term:
\begin{equation}
\label{equ:con4}
L_{Sym} = \sum_i^{\lfloor K / 2 \rfloor}{\left[W^L\left(i\right) - W^L\left(K - i\right)\right]^2}
+
\sum_i^{\lfloor K / 2 \rfloor}{\left[W^H\left(i\right) - W^H\left(K - i\right)\right]^2}.
\end{equation}


The four constraints in Eqs.~\eqref{equ:con1}-\eqref{equ:con4} jointly ensure: (1) the sum of low-pass filter weights to be $\sqrt 2$, (2) the sum of high-pass filter weights to be $0$, (3) the reversibility of pooling, and (4) the symmetric of the filter. We sum up all three terms in combination:
\begin{equation}
L_{Wavelet} = (L_{Low} + L_{High} + L_{Reverse} + L_{Sym}).
\end{equation}
For training the CNN classification model, we use cross-entropy loss which is commonly used in the classification task,
\begin{equation}
\label{equ:CELoss}
L_{CE} =  \sum_b^B{\left[\hat{y_b}\log(y_b)
+ 
\left(1 - \hat{y_b}\right)\log(1 - y_b) \right]},
\end{equation}
where $B$ denote the batch size, and $\hat{y_b}$ and $y_b$ denote the ground-truth label and model predicted probability, respectively.  The final total loss used for training is the sum of the cross-entropy loss $L_{Wavelet}$ and the constraint terms $L_{Wavelet}$:
\begin{equation}
\label{equ:regular}
L_{total} = L_{CE} + L_{Wavelet}.
\end{equation}



\subsection{Energy-based Attention Learning}
\label{sec:energy:att}

\begin{figure}[t]
\centerline{
\includegraphics[width=0.85\linewidth]{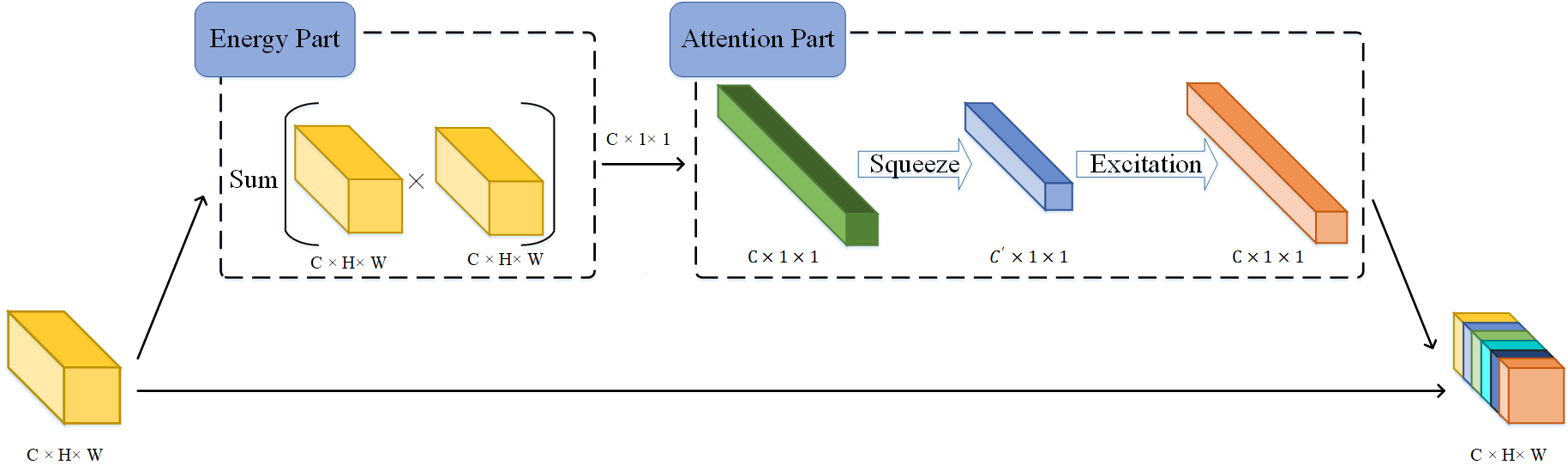}
}
\caption{
The {\bf energy attention mechanism} for feature enhancement consists of two parts. 
The ``Energy'' part calculates feature energy in each channel as the sum of squared feature values.
The ``Attention'' part learns which channel is important via squeeze-and-excitation~\cite{SENet:CVPR2018}.
}
\label{fig:EnergyAttention}
\end{figure}

To finish feature aggregation from the discrete wavelet pooling in $\S$~\ref{sec:dw:pool} so that features can be better preserved, 
we perform an energy based attention learning that can enhance the extracted feature with crucial information. Figure~\ref{fig:EnergyAttention} illustrates such energy-based attention learning mechanism. 
In contrast to the use of average in computing the feature map, the energy design aims to keep high activations. In standard average pooling, such high activations will be smoothed by the average operation when it surrounds by low activation, which is a drawback that leads to the missing of crucial information after pooling. 

In our energy-based design for pooling, we introduce a square operation in energy calculation, such that low activations that are possibly with negative values can become positive after taking square. This way, the energy-based learning can possibly attend to low input activations during the training. 
This design can essentially resolve the difficulty of learning important low activation signals, while preventing the high activations to be smoothed by low activations.


{\bf Energy definition.}
Given an input feature map $X$ of size $W \times H$, let $X^c$ denote the feature map of the $c^{th}$ feature channel. The energy $E^c$ of the feature map $X^c$ for channel $c$ is calculated as the sum of the square of all feature values in the channel:
\begin{equation}
E^c = \sum_x^W\sum_y^H \left( X_{(x,y)}^c \right)^2.
\label{eq:energy}
\end{equation}
Note that the output $Y$ of feature maps ( Eqs.\eqref{eq:LowPassPoolinginX}-\eqref{eq:HHPooling}) at current layer will be the input $X$ of next layer.  The energy mechanism reinforces all the activations no mater it responses high or low to the filter. To eliminate the impact of image brightness (whose value is usually greater than other features), we perform batch normalization before calculating the energy.

{\bf Attention calculation.}
In contrast to the simple use of global average to summarize the feature information, we use the energy calculated from Eq.~\eqref{eq:energy} to represent the channel information. Our approach is inspired from the Squeeze-and-Excitation Net (SENet)~\cite{SENet:CVPR2018}. Let $SE$ denote the Squeeze-and-Excitation without average pooling.
The Squeeze-and-Excitation with energy $S_{e}$ is calculated as:
\begin{equation}
S_{e} = SE \left( E^1,..., E^c,...,E^C \right), 
\end{equation}
where $C$ is the total number of channels. 
The feature channels $\{ X^c, c=1, ...C \}$ can then be weighted based on $S_{e}$.



\subsection{LDW-Pooling for Residual Net}
\begin{figure}[t]
\centerline{
\includegraphics[width=0.75\textwidth]{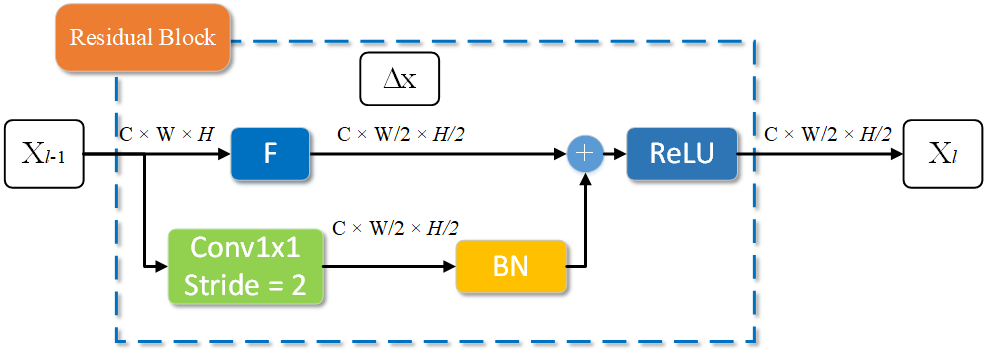}
}
\caption{{\bf Original Bottleneck.} $BN$ and $ReLU$ denote to batch normalize and ReLU function. With several convolutions to extract the feature to get $\Delta x$. The residual branch simply get features from $X_{t-1}$ with $1\times 1$ convolution with 2-stride if needed.}
\label{fig:ResBlock}
\end{figure}

\begin{figure}[t]
\centerline{
\includegraphics[width=0.9\textwidth]{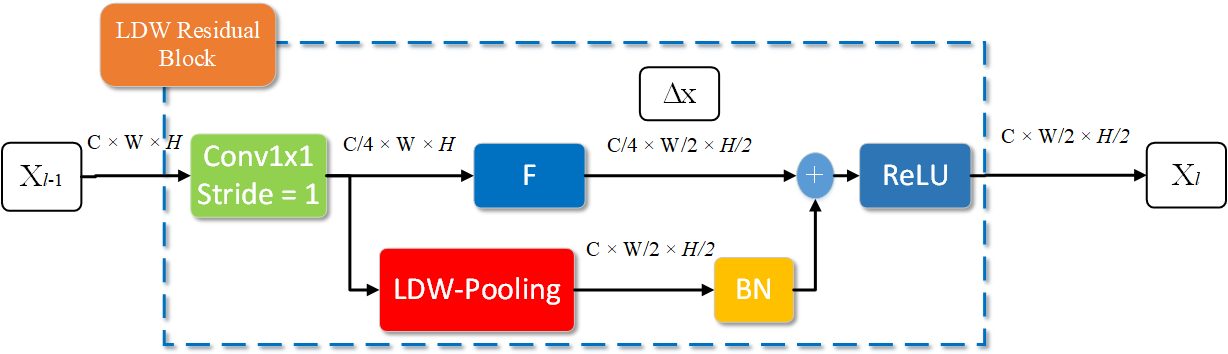}
}
\caption{{\bf LDW-Pooling in Bottleneck.} With LDW-Pooling on $X_{t-1}$ at residual branch. We slightly change the output of $Conv1\times1$ with the channel number equal to the input channel. This makes the output channel of LDW-Pooling correspond to the original output.}
\label{fig:Res:LDWPool}
\end{figure}

We describe how to apply LDW-Pooling to the widely used Residual Net~\cite{residual:cvpr2016}.
In Residual Net, the down-sample operators including max pooling and 2-stride convolution, as in Figure~\ref{fig:ResBlock}.  The time complexity to perform the residual block is ${\cal O} (\frac{1}{4} CK^2WH)$, where the kernel size for convolution is $K \times K$.

To apply LDW-pooling to the Residual Net, we first use a  $1 \times 1$ convolution with stride 2 to reduce the number of channels from $C$ to $\frac{C}{4}$. Then, the time complexity to perform the residual block is thus ${\cal O} (\frac{1}{16} CK^2WH)$. Figure~\ref{fig:Res:LDWPool} shows the details of how our LDW-pooling is applied to the bottleneck of Residual Net.  The LDW-pooling is added to the skip connection.  The number of feature channels fed to the LDW-pooling is $C/4$ and becomes $C$ after pooling.  After integration, the output dimension of this bottleneck is $C\times \frac{W}{2} \times \frac{H}{2}$. The LDW-Pooling performs 2-D convolution by decomposing it into two 2 runs of 1-D convolutions and results in the time complexity  ${\cal O} (\frac{1}{2} CKWH)$. It is thus more efficient than the original design of Residual Net.
LDW-Pooling can be easily integrated with different backbones in a similar way to leverage such efficient design.
\begin{table}[hbt!]
\caption{Ablation study of our LDW-pool on CIFAR-10~\cite{CIFAR-10:dataset} and CIFAR-100~\cite{CIFAR-100:dataset}}
\label{tab:Cifar10and100}
\centering
\begin{tabular}{cll|cll}
\hline
\multicolumn{3}{c|}{CIFAR-10}                                             & \multicolumn{3}{c}{CIFAR-100}                                      \\ \hline
\multicolumn{1}{l}{Model}    & Pooling                        & Mean Test Error       & \multicolumn{1}{l}{Model}    & Pooling                        & Mean Test Error \\ \hline
\multirow{2}{*}{ResNet50}    & \multicolumn{1}{l|}{Original}  &  8.72\%  & \multirow{2}{*}{ResNet50}    & \multicolumn{1}{l|}{Original}  & 31.89\%  \\ \cline{2-3} \cline{5-6} 
                             & \multicolumn{1}{l|}{Ours}      &  \textbf{7.92}\%  &                              & \multicolumn{1}{l|}{Ours}      & \textbf{29.75}\% \\ \hline
\multirow{2}{*}{VGG13}       & \multicolumn{1}{l|}{Original}  & 9.63\%   & \multirow{2}{*}{VGG13}       & \multicolumn{1}{l|}{Original}  & 32.06\% \\ \cline{2-3} \cline{5-6} 
                             & \multicolumn{1}{l|}{Ours}      & \textbf{9.29}\%   &                              & \multicolumn{1}{l|}{Ours}      & \textbf{30.50}\% \\ \hline
\multirow{2}{*}{VGG16}       & \multicolumn{1}{l|}{Original}  & 8.48\%   & \multirow{2}{*}{VGG16}       & \multicolumn{1}{l|}{Original}  & 30.73\% \\ \cline{2-3} \cline{5-6} 
                             & \multicolumn{1}{l|}{Ours}      & \textbf{7.94}\%   &                              & \multicolumn{1}{l|}{Ours}      & \textbf{29.91}\% \\ \hline
\end{tabular}%
\end{table}


\section{Experimental Results}
\label{sec:results}

{\bf Implementation details.}
We trained several CNN models (ResNet and VGG with various depth) with LDW-Pooling with batch size $64$ for experimental evaluation. The learning rate starts from $10^{-4}$ with a decrease by the rate of $0.1$ for every $100$ epochs. Each model is trained for $400$ epoch using the Adam optimizer with weight decay $10^{-4}$. 

\subsection{Results}
\textbf{Results on CIFAR-10.}
\label{sec:cifar10}
The dataset consists of 50,000 training data and 10,000 testing data in 10 classes of low resolution ($28\times{28}$), so we perform data augmentation on training set by re-scaling the images into $300\times{300}$ then random cropping into $224\times224$, and random flipping horizontally or vertically. For testing, we simply re-size the images into $224\times224$. 
As shown in Table~\ref{tab:Cifar10and100} (left) our proposed pooling surpass three popular backbones.


\textbf{Results on CIFAR-100.}
\label{sec:cifar100}
The CIFAR-100 dataset is similar to CIFAR-10, except that the number of the class is $100$ classes and $600$ images per class. There are $500$ and $100$ images per class for the training and testing set, respectively. We perform the same pre-processing as in CIFAR-10. As shown in Table~\ref{tab:Cifar10and100} (right), that LDW-Pooling has sigfinicant performance gain over all the baselines due to the property of feature preserving. 

\textbf{Results on PASCAL-VOC12.}
\label{sec:pascal-voc12}
The PASCAL-VOC12 is a semantic segmentation dataset that contains 20 foreground object classes and one background class. The augmented version of PASCAL-VOC12 which contains 10582 training images and 1449 validation images is used for training and validation. We consider the basic UNet model which downsamples with maximum pooling and upsamples with de-convolution as the benchmark. We change the maximum pooling and de-convolution with LDW-Pooling and reverse LDW-Pooling for the comparison. The performance is measured in terms of pixel mean-intersection-over-union(mIoU) across 21 classes as shown in Table~\ref{tab:UNet}. The reversibility of the LDW pooling method makes better segmentation results gained than the original Unet.  Figure~\ref{fig:LDWpool}(b) shows the reconstructed result after LDW-Pooling decomposition and recovering by UNet with PSNR=33.1.

\textbf{Results on ImageNet.}
\label{sec:imagenet}
ImageNet consists of 1,000 categories ranging from 732 to 1300 images per class in the training set and 50 images per class in the validation set. We performed the same pre-processing as done in Section~\ref{sec:cifar10} for CIFAR-10. Table~\ref{tab:ImageNet} shows the comparison of LDW-Pooling with other state-of-the-art pooling methods. For all the categories except ResNet 18, our method outperforms other methods. For ResNet18, fewer convolution layers might cause few semantic features to be extracted for LDW-pooling.   

\subsection{Ablation Study}
To show the effectiveness of our method, we conduct ablation study on several settings.

\textbf{LDW-Pooling and Energy Attention on classification} is reported with CIFAR-10/100 in Table~\ref{tab:ablation_ea}. It is clear that the accuracy improvement is majorly caused by the LDW-Pooling technique rather than energy attention.

\textbf{LDW-Pooling vs i-RevNet~\cite{i-revnet:ICLR2018} 
Table~\ref{tab:comparison_ps} shows that comparing with SoTA reversible pooling method, $i.e.$, i-RevNet~\cite{i-revnet:ICLR2018}. Our LDW-pooling method can gain better classification accuracy on both CIFAR-10 and CIFAR-100 dataset.  Since i-RevNet~\cite{i-revnet:ICLR2018} focuses only on the design of filters with good reversibility rather than good feature extraction.}  

\textbf{Pretrained filter weights and $L_{wavelet}$} is verified in
Table~\ref{tab:ablation_Lwavelet}. The pretrained weights were obtained by randomly generating filters which satisfy wavelet constraints before training.  Clearly, better pooling filters for target tasks can be derived via the loss $L_{wavelet}$. More weights on $L_{Wavelet}$ will enhance the reversibility with higher PSNR if the target task is "reconstruction" or "segmentation".  For the classification task, $L_{CE}$ is a necessary term.  Table \ref{tab:ablation_Lwavelet} shows the accuracy of classification can be further improved if the term  $L_{Wavelet}$ is added.  If  the loss  $L_{Wavelet}$ is added more, more classification accuracy can be gained.  However, when the reversibility is higher, the effect of $L_{CE}$ will decrease and result in accuracy degradation.   Thus, the weights for $L_{CE}$ and $L_{Wavelet}$  are equally set (see Eq.(\ref{equ:regular})).  The "pretrain" option means initial weights are obtained by randomly generating filters which satisfy wavelet constraints before training.
\begin{table*}
\caption{Comparisons among LDW-Pooling and SoTA pooling methods based on ImageNet~\cite{ImageNet:CVPR2009}. Symbol $\star$ denotes the re-evaluated scores by running source codes from original authors.  Other scores were cited from \cite{LiftPooling:ICLR2021}.}
\label{tab:ImageNet}
\centering
\begin{tabular}{lllllll}
\hline
              & \multicolumn{2}{c}{ResNet18}    & \multicolumn{2}{c}{Renst50} & \multicolumn{2}{c}{MobileNet-V2} \\ \cline{2-7} 
              & Top-1            & Top-5        & Top-1        & Top-5        & Top-1           & Top-5          \\ \hline
Skip~\cite{Striving:ICLR2015}        & 30.22            & 10.23        & 24.31        & 7.34         & 28.66           & 9.70           \\
MaxPool       & 28.60            & 9.77         & 24.26        & 7.22         & 28.65           & 9.82           \\
AveragePool   & 28.03            & 9.55         & 24.40        & 7.35         & 28.32           & 9.72           \\ \hline
S3Pool~\cite{s3pool:CVPR2017}    & 33.91            & 13.09        & 27.98        & 9.34         & 40.56           & 17.91          \\
WaveletPool~\cite{WaveletPooling:ICLR2018}   & 30.33            & 10.82        & 24.43        & 7.36         & 29.27           & 10.26          \\
BlurPool~\cite{Making:ICML2019}$\star$     & 29.88            & 10.58        & 24.60        & 7.73         & 30.58           & 11.26          \\
DPP~\cite{detailpreserving:CVPR2018}$\star$        & 29.12            & 10.21        & 24.62        & 7.49         & 29.85           & 10.53          \\
SpectralPool~\cite{Spectral_nips2015}  & 28.69            & 9.87         & 24.81        & 7.57         & 33.38           & 12.56          \\
GatedPool~\cite{Generalizing:AISTATS2016}    & 27.78            & 9.44         & 23.79        & 7.06         & 28.94           & 9.90           \\
MixedPool~\cite{Generalizing:AISTATS2016}     & 27.76            & 9.50         & 24.08        & 7.32         & 29.00           & 9.97           \\
GGFGP~\cite{Gobal:ICCV2019}$\star$       & 26.88            & 8.66         & 22.76        & 6.34         & 28.42           & 9.59           \\
GaussPool~\cite{Gaussian:nips2019a}$\star$     & 26.58            & 8.86         & 22.95        & 6.30         & 27.13           & 8.92           \\
LiftDownPool~\cite{LiftDownPool:ICLR2021}  & \textbf{25.80}            & 8.14         & 22.36        & 6.11         & 26.09           & 8.22           \\ \hline
LDW-Pool & 26.10  & \textbf{8.01} &  \textbf{22.10}  & \textbf{5.88}  &  \textbf{25.97}  &  \textbf{7.98 }  \\ \hline
\end{tabular}%
\end{table*}

\begin{table}[h!]
\caption{Ablation study of LDW-Pooling and Energy Attention.}
\label{tab:ablation_ea}
\centering{%
\begin{tabular}{cclll}
\hline
Model                     & LDW-Pooling          & \multicolumn{1}{c}{Energy Attention} & \multicolumn{1}{c}{\begin{tabular}[c]{@{}c@{}}CIFAR-10\\
Mean Test Error\end{tabular}} & \multicolumn{1}{c}{\begin{tabular}[c]{@{}c@{}}CIFAR-100\\ Mean Test Error\end{tabular}} \\ \hline
\multirow{3}{*}{ResNet50} & V                    & \multicolumn{1}{c}{V}  & \multicolumn{1}{c}{7.92\%}  & \multicolumn{1}{c}{29.75\%} \\
                          & V                    &                        & \multicolumn{1}{c}{7.98\%}  & \multicolumn{1}{c}{29.83\%} \\
                          & \multicolumn{1}{l}{} &                        & \multicolumn{1}{c}{8.72\%}  & \multicolumn{1}{c}{31.89\%} \\ \hline
\end{tabular}%
}
\end{table}

\begin{table}[h!]
\caption{Ablation study of reversibility with UNet on VOC12 dataset.}
\label{tab:UNet}
\centering{%
\begin{tabular}{cclll}
\cline{1-3}
\multicolumn{3}{c}{VOC12   (21 classes)} &  &  \\ \cline{1-3}
Model                 & Pooling                                                                           & mIoU  &   &  \\ \cline{1-3}
\multirow{2}{*}{Unet} & Original & 47.83 &   &  \\ \cline{2-3}
                      & Ours                                                                            & 49.02 &   &  \\ \cline{1-3}
\end{tabular}%
}
\end{table}

\begin{table}[h!]
\caption{Comparisons between our method and i-RevNet\cite{i-revnet:ICLR2018}.}
\label{tab:comparison_ps}
\centering{%
\begin{tabular}{cccc}
\hline
\multicolumn{1}{l}{Model} & Pooling & \multicolumn{1}{c}{\begin{tabular}[c]{@{}c@{}}CIFAR-10\\      Mean Test Error\end{tabular}} & \multicolumn{1}{c}{\begin{tabular}[c]{@{}c@{}}CIFAR-100\\      Mean Test Error\end{tabular}} \\ \hline
\multirow{2}{*}{ResNet50} & Our     & 7.92\% & 29.75\%  \\
                          & PS      & 8.77\% & 32.02\%  \\ \hline
\end{tabular}%
}
\end{table}

\begin{table}[h!]
\caption{Ablation study of the effect of $L_{Wavelet}$.}
\label{tab:ablation_Lwavelet}
\centering{%
\begin{tabular}{ccccc}
\cline{1-3}
Pretrain                     & L\_wavelet              & Mean Test Error     &  &  \\ \cline{1-3}
\multicolumn{1}{c}{V}       & \multicolumn{1}{c}{V}   & 7.92\% &  &  \\
\multicolumn{1}{c}{V}       &                         & 7.98\%   &  &  \\
                            & \multicolumn{1}{c}{V}   & 9.13\%   &  &  \\
                            &                         & 9.94\%   &  &  \\ \cline{1-3}
\end{tabular}%
}
\end{table}



\section{Conclusion}

This paper presented the LDW-Pooling, an effective learnable pooling method that can swap and replace the widely-used pooling methods in CNN with many desirable advantages. The discrete wavelet transform properties ensure the invertibility of the extracted features after pooling. The weights of the wavelet kernel can be learned from training data via an energy-based attention mechanism to better perform in each application-dependent scenario. The 1-D linear complexity is computationally efficient when compared with other wavelet-based pooling methods. Experimental evaluations show that LDW-Pooling outperforms other wavelet-based pooling approaches as well as standard pooling methods. 

{\bf Future work} includes further application and evaluation of the LDW-Pooling on other computer vision tasks such as object detection with segmentation, as well as application to non-vision tasks including NLP and others.

\clearpage

\bibliographystyle{unsrt}  
\bibliography{pool}  

\begin{thebibliography}{10}

\bibitem{Making:ICML2019}
Richard Zhang.
\newblock Making convolutional networks shift-invariant again.
\newblock In {\em ICML}, 2019.

\bibitem{LiftPooling:ICLR2021}
Jiaojiao Zhao and Cees~G.M. Snoek.
\newblock Liftpool : Bidirectional convnet pooling.
\newblock In {\em ICLR}, 2021.

\bibitem{WaveletPooling:ICLR2018}
Travis Williams and Robert Li.
\newblock Wavelet pooling for convolutional neural networks.
\newblock In {\em ICLR}, 2018.

\bibitem{Stochastic:ICLR2013}
Matthew~D. Zeiler and Rob Fergus.
\newblock Stochastic pooling for regularization of deep convolutional neural
  networks.
\newblock In {\em ICLR}, 2013.

\bibitem{Lip:ICCV2019}
Ziteng Gao, Limin Wang, and Gangshan Wu.
\newblock {LIP}: Local importance-based pooling.
\newblock In {\em ICCV}, 2019.

\bibitem{Adaptive:WACV2020}
Shin Fujieda, Flavio~Prieto Ortiz, Kohei Takayama, and Kohei Takayama.
\newblock Deep adaptive wavelet network.
\newblock In {\em WACV}, 2020.

\bibitem{Mixedpooling:Springer2014}
Dingjun Yu, Hanli Wang, Peiqiu Chen, and Zhihua Wei.
\newblock Mixed pooling for convolutional neural networks.
\newblock In {\em Rough Sets and Knowledge Technology}, 2014.

\bibitem{detailpreserving:CVPR2018}
Faraz Saeedan, Nicolas Weber, Michael Goesele, and Stefan Roth.
\newblock Detail-preserving pooling in deep networks.
\newblock In {\em CVPR}, 2018.

\bibitem{Generalizing:AISTATS2016}
Chen-Yu Lee, Patrick~W Gallagher, and Zhuowen Tu.
\newblock Generalizing pooling functions in convolutional neural networks:
  Mixed, gated, and tree.
\newblock In {\em AISTATS}, 2016.

\bibitem{s3pool:CVPR2017}
Shuangfei Zhai, Hui Wu, Abhishek Kumar, Yu~Cheng, Yongxi Lu, Zhongfei~(Mark)
  Zhang, and Rogerio Feris.
\newblock {S3Pool}: Pooling with stochastic spatial sampling.
\newblock In {\em CVPR}, 2017.

\bibitem{Gaussian:nips2019a}
Kobayashi Takumi.
\newblock Gaussian-based pooling for convolutional neural networks.
\newblock In {\em Advances in Neural Information Processing Systems (NeurIPS)},
  volume~32, 2019.

\bibitem{Gobal:ICCV2019}
Takumi Kobayashi.
\newblock Global feature guided local pooling.
\newblock In {\em ICCV}, 2019.

\bibitem{Scaling:ICCV2017}
Edouard Oyallon, Eugene Belilovsky, and Sergey Zagoruyko.
\newblock Scaling the scattering transform: Deep hybrid networks.
\newblock In {\em ICCV}, 2017.

\bibitem{WaveletConv:CoRR2018}
Shin Fujieda, Kohei Takayama, and Toshiya Hachisuka.
\newblock Wavelet convolutional neural networks.
\newblock {\em CoRR}, abs/1805.08620, 2018.

\bibitem{Liftingscheme:SIAM1998}
Wim Sweldens.
\newblock The lifting scheme: A construction of second generation wavelets.
\newblock In {\em SIAM Journal on Mathematical Analysis}, 1998.

\bibitem{SENet:CVPR2018}
Hu, Jie, Shen, Li, and Gang Sun.
\newblock Squeeze-and-excitation networks.
\newblock In {\em CVPR}, 2018.

\bibitem{residual:cvpr2016}
Kaiming He, Xiangyu Zhang, Shaoqing Ren, and Jian Sun.
\newblock Deep residual learning for image recognition.
\newblock In {\em CVPR}, June 2016.

\bibitem{CIFAR-10:dataset}
Alex Krizhevsky, Vinod Nair, and Geoffrey Hinton.
\newblock {CIFAR-10} dataset, 2009.
\newblock \url{http://www.cs.toronto.edu/~kriz/cifar.html}.

\bibitem{CIFAR-100:dataset}
Alex Krizhevsky, Vinod Nair, and Geoffrey Hinton.
\newblock Cifar-100 (canadian institute for advanced research).
\newblock 2009.

\bibitem{i-revnet:ICLR2018}
Jorn-Henrik Jacobsen, Arnold Smeulders, and Edouard Oyallon.
\newblock i-revnet: Deep invertible networks.
\newblock In {\em ICLR}, 2018.

\bibitem{ImageNet:CVPR2009}
J.~Deng, W.~Dong, R.~Socher, L.-J. Li, K.~Li, and L.~Fei-Fei.
\newblock {ImageNet}: A large-scale hierarchical image database.
\newblock In {\em CVPR}, 2009.

\bibitem{Striving:ICLR2015}
Jost~Tobias Springenberg, Alexey Dosovitskiy, Thomas Brox, and Martin
  Riedmiller.
\newblock Striving for simplicity:the all convolutional net.
\newblock In {\em ICLR}, 2015.

\bibitem{Spectral_nips2015}
Oren Rippel, Jasper Snoek, and Ryan~P Adams.
\newblock Spectral representations for convolutional neural networks.
\newblock In {\em Advances in Neural Information Processing Systems (NeurIPS)},
  volume~28, 2015.

\bibitem{LiftDownPool:ICLR2021}
Jiaojiao Zhao and Cees G.~M. Snoek.
\newblock Liftpool: Bidirectional convnet pooling.
\newblock In {\em ICLR}, 2021.

\end{thebibliography}

\end{document}